\documentclass[10pt,twocolumn,letterpaper]{article}

\usepackage{cvpr}              % To produce the CAMERA-READY version
\usepackage{graphicx}
\usepackage{amsmath}
\usepackage{amssymb}
\usepackage{booktabs}

\usepackage[pagebackref,breaklinks,colorlinks]{hyperref}

\usepackage[capitalize]{cleveref}
\crefname{section}{Sec.}{Secs.}
\Crefname{section}{Section}{Sections}
\Crefname{table}{Table}{Tables}
\crefname{table}{Tab.}{Tabs.}

\def\cvprPaperID{*****} % *** Enter the CVPR Paper ID here
\def\confName{CVPR}
\def\confYear{2022}

\begin{document}

\title{FaceChain-FACT: Face Adapter with Decoupled Training for Identity-preserved Personalization}
\author{
Cheng Yu\footnote{Equal contribution}, Haoyu Xie\footnotemark[\value{footnote}], Lei Shang, Yang Liu, Jun Dan, Liefeng Bo, Baigui Sun\footnote{Corresponding author} \\
  Alibaba Group, FaceChain Community \\
  \texttt{\{yucheng.yu, xiehaoyu.xhy\}@alibaba-inc.com} \\
% First Author\\
% \and
}
\maketitle

%%%%%%%%%%%%%%%%%%%%%%%%%%%%%%%%%%%% ABSTRACT %%%%%%%%%%%%%%%%%%%%%%%%%%%%%%%%%%%%
\begin{abstract}
\label{sec:abstract}
Recently, text-to-image diffusion models have gained widespread attention in the community because of their high-fidelity image generation capability.
Specifically, in the field of human-centric personalized image generation, the adapter-based method obtains the ability to customize and generate portraits by text-to-image training on facial data.
This allows for identity-preserved personalization without additional fine-tuning in inference.
Although there are improvements in efficiency and fidelity, there is often a significant performance decrease in test following ability, controllability, and diversity of generated faces compared to the base model.
In this paper, we analyze that the performance degradation is attributed to the failure to decouple identity features from other attributes during extraction, as well as the failure to decouple the portrait generation training from the overall generation task.
To address these issues, we propose the Face Adapter with deCoupled Training (FACT) framework, focusing on both model architecture and training strategy. 
To decouple identity features from others, we leverage a transformer-based face-export encoder and harness fine-grained identity features.
To decouple the portrait generation training, we propose Face Adapting Increment Regularization~(FAIR), which effectively constrains the effect of face adapters on the facial region, preserving the generative ability of the base model.
Additionally, we incorporate a face condition drop and shuffle mechanism, combined with curriculum learning, to enhance facial controllability and diversity.
As a result, FACT solely learns identity preservation from training data, thereby minimizing the impact on the original text-to-image capabilities of the base model.
Extensive experiments show that FACT has both controllability and fidelity in both text-to-image generation and inpainting solutions for portrait generation.
Our codes are available in \url{https://github.com/modelscope/facechain}.

\end{abstract}

%%%%%%%%%%%%%%%%%%%%%%%%%%%%%%%% INTRODUCTIOIN %%%%%%%%%%%%%%%%%%%%%%%%%%%%%%
\section{Introduction}
\label{sec:intro}
As text-to-image technology, such as GLIDE~\cite{glide}, DALL-E~\cite{dalle2}, Imagen~\cite{imagen}, and Stable Diffusion~\cite{sd}, continues to evolve, users have increasingly demanded higher capabilities from these advancements.
How to integrate customized portraits into generated images has recently received widespread attention from the community.
Personalized image generation technology, \eg DreamBooth~\cite{dreambooth}, Textual Inversion~\cite{textual_inversion}, and LORA~\cite{lora} has paved a way.
Users can provide images that contain customized portraits, and the text-to-image model learns the characteristics of these portraits through fine-tuning, generating images around these portraits.
However, personalized methods based on fine-tuning have inherent limitations that hinder their ability to satisfy the demands in certain scenarios.
First, without domain expert level encoding of the customized faces, these methods will lose lots of detailed identity information during generation, resulting in reduced fidelity.
Then, fine-tuning requires a significant amount of training costs, such as time and computing resources, which are difficult to meet user needs in many practical scenarios.

To address the aforementioned issues, a series of methods~\cite{ipadapter,photomaker,caphuman,instantid} utilize the generalization properties of facial identity features and use customized faces as additional conditions for the Stable Diffusion model to generate personalized portraits instantly.
Specifically, they treats facial embedding of input portrait images as a condition parallel with textual embeddings for T2I generation.
They insert an adapter architecture~\cite{ipadapter} into the middle of the Stable Diffusion's U-Net blocks to integrate facial embeddings, and train the adapter while freezing other modules.
By text-to-image training on the large amount of portrait data, the diffusion model learns the identity feature of the faces, and then performs personalized generation using a single image with a single forward pass without additional fine-tuning.

Despite the efficiency, there are two critical issues encountered in the process of integrating portrait generation functionality.
a) Firstly, the vividness and fidelity of faces are not satisfactory, resulting in generated portraits that look like copies with reduced controllability and diversity.
As shown on the right side of Fig.~\ref{fig:intro}, under detailed text prompts, the quality of the generated portrait will significantly decrease and artifacts will appear.
% In addition, other characteristics such as clothing and expressions are also retrained in addition to identity.
b) Secondly, compared to the basic diffusion model, there is a significant performance decline in the text-to-image ability.
As shown on the left side of Fig.~\ref{fig:intro}, there is a decrease in quality and complexity outside the facial region, such as monotonous backgrounds and distorted clothing.
Besides, there are also issues in the generated images such as disrupted spatial continuity between faces and bodies, as well as the absence of glasses mentioned in the prompt.

In this paper, we rethink the identity-preserved personalization task and divide it into two processes: separating identity features from portrait images and integrating them into the generation process.
From this view, the causes of the two issues mentioned above can be attributed to two aspects respectively:
a) The failure to decouple identity features from others during the process of separating identity features from portrait images results in other coupled features involved hindering the generation of portraits.
b) The failure to decouple the portrait generation training from the overall generation task leads to a decline in the ability of original text-to-image generation.

Therefore, we propose Face Adapter with deCoupled Training~(FACT) based on Stable Diffusion.
It focuses on both model architecture and training strategy to achieve a clear decoupling of identity features and impose stronger constraints on portrait generation tasks.
Specifically for decoupling the identity feature, we mask the region outside the face, effectively eliminating other distractions in the photograph.
We harness fine-grained identity features from a transformer-based face recognition model.
These features can better decouple information unrelated to identity and contain richer details of facial features.
In addition, we insert a sequential Gated Self-Attention~(GSA) module on U-Net to adapt independently to facial features, reducing the interference of textual embeddings compared to parallel cross attention.
For decoupling the portrait generation training, we introduce the "latent face adapting" task to make the adapter focus on adapting the facial identity of the latent in denoising process.
Specifically, we propose Face Adapting Increment Regularization~(FAIR), which limits the effect of face adapters on the face region by restricting the increment through GSA modules in deep layers of the U-Net.
Besides, to increase facial controllability and diversity, we design a curriculum learning with face condition drop and shuffle.

We conduct extensive experiments to verify the effectiveness of the proposed FACT.
As shown in the Fig.~\ref{fig:intro}, our proposed FACT preserves the text-to-image ability of the original model, which is manifested in diverse backgrounds and appropriate clothing.
And FACT outperforms existing adapter-based personalization methods in terms of identity similarity, text following ability, and facial controllability and diversity.
Besides, FACT can seamlessly integrate with common fine-tuning models of Stable Diffusion like LORA~\cite{lora} and ControlNet~\cite{controlnet} without compromising on generation performance.

% To sum up, our contributions are as follow:
% \begin{itemize}
%     \item
%     \item
%     \item
% \end{itemize}

%  

%%%%%%%%%%%%%%%%%%%%%%%%%%%%%%%% Related Work %%%%%%%%%%%%%%%%%%%%%%%%%%%%%%
\section{Related Work}
\label{sec:relat}
\begin{figure*}[t]
  \centering
  \includegraphics[width=1.0\linewidth]{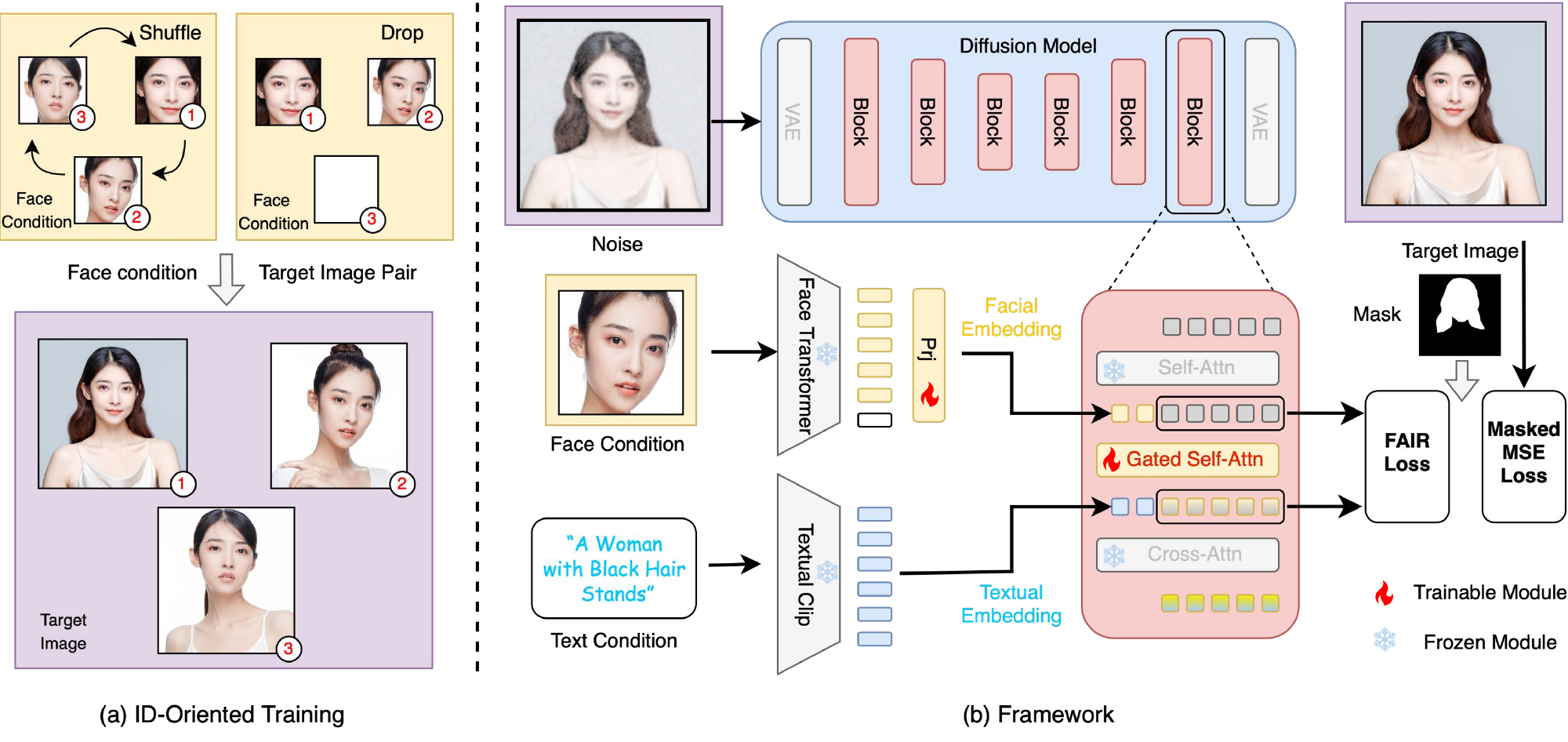}
  \caption{\textbf{Overview of the proposed (a) ID-Oriented training and (b) framework of FACT.} 
  (a) We conduct random face condition drop and shuffling from the same identity, corresponding to the "unpaired identities" and "paired identities + unpaired faces".
  (b) Firstly, we leverage a transformer-based face expert encoder to extract fine-grained identity features from the input portrait image.
  Subsequently, we insert SFAGS to integrate these identity features to the diffusion process.
  Within the U-Net block, we employ a FAIR loss to constrain the impact of the proposed SFAGS specifically within the face region.
  Finally, combined with a mask MSE loss, we fine-tune the SFAGS to integrate the capability of identity-preserved personalization into the base model.
  % We perform \textbf{identity-guided face denoising} instead of \textbf{face-guided image denoising} for decoupled training. We insert sequential face adapters between the self-attention block and the cross-attention block to perform ``\textbf{latent face adapting}'' contrained by Face Adapting Increment Regularization (FAIR) on the latents before text guidance, thus avoid interfering face condition with text information. For an image to denoise, the identity condition comes from a face image with the same identity by face shuffling, and is extracted by a transformer-based feature extracter.
  }
  \label{fig:overview}
\end{figure*}

\subsection{Text-to-Image Diffusion Models}
With the rapid progress of large-scale datasets like LAION-5B~\cite{LAION-5B} and powerful computation supports, large diffusion models~\cite{ddpm, ddim} have made remarkable advancement in text-to-image generation~\cite{imagic, saharia2022photorealistictexttoimagediffusionmodels, sd} due to their scalability without mode collapse of adversarial training~\cite{gan}, and attract widespread attention in recent years.
Methods like GLIDE~\cite{glide}, DALL-E~\cite{dalle2} and Imagen~\cite{imagen} directly perform denoising in the pixel space, while Stable Diffusion series~\cite{sd, sdxl} perform it in the latent space, which enables training and fine-tuning using less computation resources without significantly affecting overall generation performance.
As such, numerous fine-tuning models including upgrades of foundation models~\cite{videosd}, controlling models~\cite{controlnet} and personalized models~\cite{dreambooth} bloom in the community of Stable Diffusion.
Our proposed FACT is also based on Stable Diffusion considering the above aspects.

\subsection{Personalization on Diffusion Models}
Although large text-to-image diffusion models can learn strong semantic priors, it is hard to realize more fine-grained control such as generating images with specific identity or style using text condition only.
Therefore, more researchers try to explore personalization methods for diffusion models to preserve the identity of a specific subject given one or a few reference images.
There are two main branches for personalization.
A series of researches, such as Textual Inversion~\cite{textialinversion}, DreamBooth~\cite{dreambooth}, and LoRA~\cite{lora} are proposed to perform additional fine-tuning during the test phase, making the diffusion model generating images similar to the reference images.
Despite the difference in trained parameters and optimizing strategies for those methods, they generally requires a significant amount of computational costs.
Other works~\cite{gligen, ipadapter} use the adapter-based architecture to the original diffusion model by adding attention to the input image.
By offline fine-tuning on the subject data, the model learns to encode the generalizable feature for identity preservation.
However, such approaches tend to overfit the entire image instead of preserving the identity of face images.
To solve this, PhotoMaker~\cite{photomaker} introduces multiple face images to get the stacked ID embedding to discern the characteristics of the identity information.
However, the identity similarity is relatively low when using only one reference image, and the image quality also drops significantly compared to its base model.
InstantID~\cite{instantid} and CapHuman~\cite{caphuman} try to couple the identity information with face spatial conditions like facial landmarks and 3D Morphable Models (3DMM)~\cite{3dmm} to enhance the face shape control.
However, they lack the ability of generating faces with various poses, and they cannot solve the drop of face editability with text conditions.
Different from the above approaches, we propose to decouple the identity information from the reference image for adapter training in terms of both model architecture and training strategy, and improve the controllability and authenticity in portrait generation without introducing extra input faces or structural conditions.

%%%%%%%%%%%%%%%%%%%%%%%%%%%%%%%% Methods %%%%%%%%%%%%%%%%%%%%%%%%%%%%%%
\section{Method}
\label{sec:method}
\subsection{Overview}

Given one single portrait image to be customized, the goal of identity-preserved personalization for a text-to-image diffusion model is to perform image generation that preserves the identity information of the input portrait image and retains other properties of the original model.
Unlike methods such as IPAdapter~\cite{ipadapter} and Photomaker~\cite{photomaker} that only consider how to integrate portraits into the generation process, our goal is to achieve better identity decoupling ability from portrait images and lossless generation ability of the diffusion model.
In order for the goals above, we propose Face Adapter with deCoupled Training (FACT).
From both aspects of model architecture and training strategy, we design an Identity Merging Module(IMM) to obtain better identity decoupling and merging capability, as well as a Face Adapting Increment Regularization(FAIR) to achieve face construction while preserving the original generation ability of the diffusion model.
Fig.~\ref{fig:overview} shows the overall framework of FACT including model architecture, training strategy, and data preprocessing. 

\subsection{Identity Merging Module}
Our Identity Merging Module~(IMM) consists of a face encoder and a sequential face adapter.
The former encodes the input portrait image to the identity feature, and then merges the identity feature into the Stable Diffusion.

\noindent\textbf{Face Encoding.} 
We consider the following two aspects for an ideal identity feature for guiding portrait generation in diffusion.
Firstly, the identity feature should align the feature space of the diffusion model for attention.
Secondly, the identity feature should have the property of intra-indentity compactness and inter-identity separation as well as containing enough information in terms of image generation.
Taking into account the above factors, we use the TransFace model~\cite{transface} for feature extraction, which is a ViT-based face recognition model to achieve better identity discrimination and compatibility with the Stable Diffusion architecture.
Compared with general image encoders such as CLIP \cite{clip}, the identity features extracted by the face expert encoder have stronger intra-identity compactness and inter-identity separation.
To obtain richer facial information, we leverage the feature representations for all visual tokens from the penultimate layer of the TransFace.
For alignment with Stable Diffusion architecture, we add a learnable lightweight projection module $Prj$ (several Transformer blocks).
Specifically, denote $PTF$ as the token extraction from the penultimate ViT layer of TransFace, then the extraction of face embedding $\mathbf{e}_{id}$ can be represented as:
\begin{equation}
\label{eq: faceencode}
    \mathbf{e}_{id} = Prj(PTF(\mathbf{x}_{face})),
\end{equation}
where $\mathbf{x}_{face}$ represents the input portrait image.

\noindent\textbf{Sequential Face Adapter with Gated Self-Attention.}
Inspired from the sequential sequential operation of face swapping methods which only changes the facial identity in the face region but keep other attributes constant, we use Sequential Face adapter with Gated Self-Attention~(SFAGS) to merge the ID information contained in face embedding $\mathbf{e}_{id}$.
We insert a new gated cross-attention layer guided by the face embeddings between the self-attention layer and the cross-attention layer of the transformer block.
We freeze the original two attention layers and train the gated cross attention layer for identity merging ability as is used in Flamingo~\cite{Flamingo} and GLIGEN~\cite{gligen}.
The forward pass of SFAGS is formulated as:
\begin{equation}
\begin{aligned}
\label{eq: SFAGS}
    \mathbf{x} &= \mathbf{x} + \alpha \cdot GSA(\mathbf{x}),\\
    GSA(\mathbf{x}) &= \text{tanh}(\gamma)\cdot TS(\text{SelfAttn}([\mathbf{x}, \mathbf{e}_{id}])),
    \end{aligned}
\end{equation}
where $\mathbf{x}$ represents visual tokens, $TS(\cdot)$ is the token selection operation that considers the visual tokens only, $GSA$ is the gated self-attention.
$\alpha$ denotes the scale of the face adapter, and $\gamma$ is a learnable parameter initialized as $0$ for stable training.
As such, the face adapter only performs face adapting on the latent before cross-attention, rather than intervening in the text embedding within cross-attention.
This approach minimizes the interference between face and text, thereby better preserving the guiding role of the text.

\subsection{Identity Preserving Training}
Our goal is to perform identity preservation training rather than tend to face reconstruction as some methods do.
This signifies that during the portrait generation process, we need to increase the variability and controllability of the portrait while maintaining its authenticity as a prerequisite.
Taking into account this, we use Face Adapting Increment Regularization~(FAIR) to make the training process focus on the face region, thus improving the authenticity.
In addition, we use a curriculum learning strategy with face condition drop and shuffle to expand the variability and controllability of portrait generation.

\begin{figure}[t]
  \centering
  \includegraphics[width=1.0\linewidth]{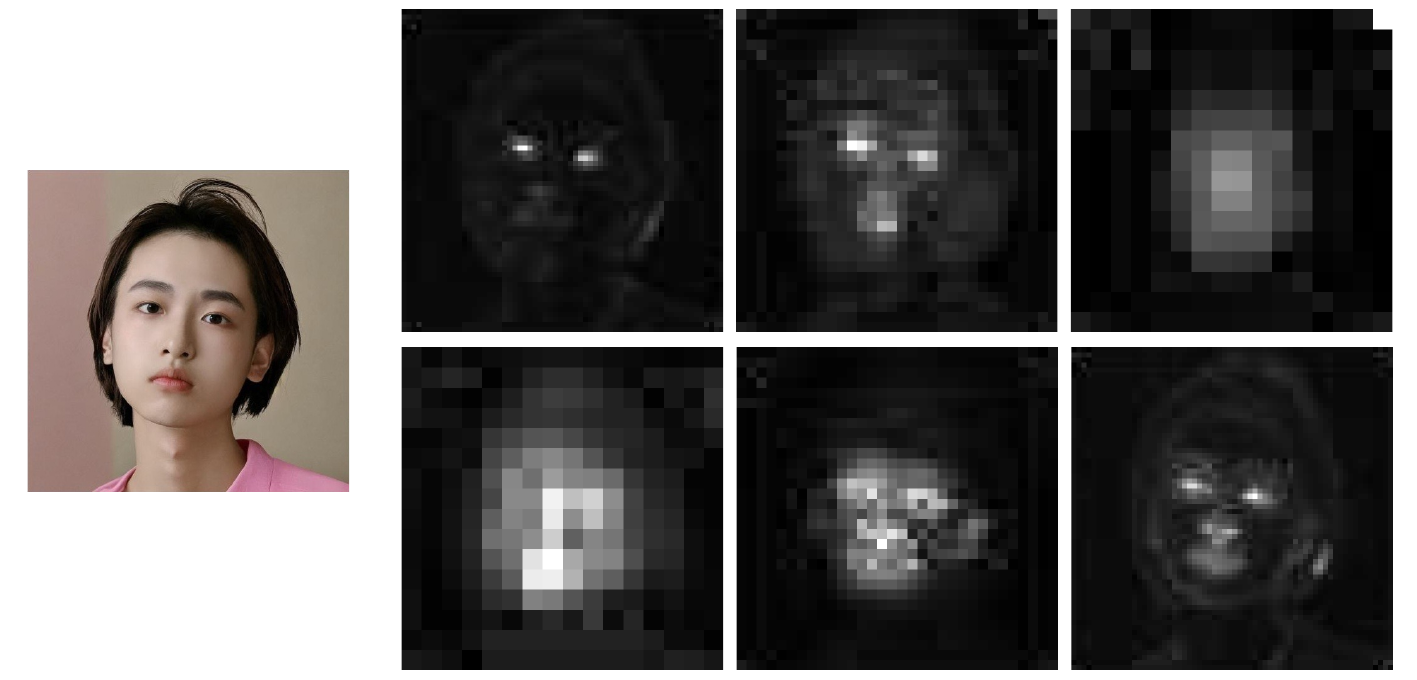}
  \caption{\textbf{Normalized face adapting increment in different depths of Transformer blocks.} The increment for each block is averaged across all timesteps in the T2I generation. The indices of the Transformer blocks are 1, 3, 5, 9, 12, 15 respectively.}
  \label{fig:fair}
\end{figure}

\noindent\textbf{Face Adapting Increment Regularization.}
We modify the training objective for the model optimization to constrain the impact of the proposed SFAGS specifically within the face region.
A naive method is face-masked diffusion loss, but it is not strong enough to constrain the impact of SFAGS by only controlling the final denoising result without directly controlling the input and output of each adapter.
Some adapter-based methods try to constrain the spatial distribution of attention deeply within the model, making it strictly focused on the facial region.
However, we have observed that the attention map does not strictly correspond to the impact of the face adapter, and features on different intermediate layers focus on different regions of the face, which is demonstrated in Figure \ref{fig:fair}.
Such a hard constraint ignores the attention difference in different parts within the face for different layers.
Taking this in consideration, benefiting from the sequential structure of the face adapter, we propose a soft method to constrain the model. 
Essentially, the impact of the sequential face adapter is the increment of the visual token $x$ in Eq.~\ref{eq: SFAGS}.
Therefore, we propose a Face Adapting Increment Regularization~(FAIR) to decrease the impact of sequential face adapter on the outside of the face region, formulated as:
\begin{equation}
    \label{eq:fair}
    \mathcal{L}_{\text{FAIR}}(\mathbf{x}) = \frac{\Vert GSA(\textbf{x}) \odot (1 - \mathbf{M}_{\textbf{x}}) \Vert}{\Vert \textbf{x} \odot (1 - \mathbf{M}_{\textbf{x}}) \Vert},
\end{equation}
where $\mathbf{M}_\mathbf{x}$ is the mask of the corresponding face mask on the latent space with the same shape of $\mathbf{x}$, $\Vert \cdot \Vert$ is the $\mathcal{L}_{2}$ norm.
As such, FAIR retrains the increment by limiting the relative variation of latent caused by the face adapter in areas outside of the face and paying attention to different areas inside the face.
Finally, combined with masked diffusion loss, the total training loss can be represented as:
\begin{equation}
\scalebox{1.0}{$\displaystyle
    \begin{aligned}
    \label{eq:loss}
    \mathcal{L} = &\mathbb{E}_{\mathbf{z}, \mathbf{e}_{text}, \mathbf{e}_{id},  \mathbf{\epsilon} \sim \mathcal{N}(\mathbf{0}, \mathbf{I}), t} \ (\lambda \sum_{\mathbf{x}} \mathcal{L}_{\text{FAIR}}(\mathbf{x}_{\mathbf{z}_t, t, \mathbf{e}_{text}, \mathbf{e}_{id}}) \\
    +&\Vert (\mathbf{\epsilon}_{\theta}(\mathbf{z}_t, t, \mathbf{e}_{text}, \mathbf{e}_{id}) - \mathbf{\epsilon}) \odot \mathbf{M}_r(p))\Vert, 
    \end{aligned}
    $}
\end{equation}
where $\mathbf{M}$ is the mask of the face, $\mathbf{z}$ is the original input image latent, $\mathbf{e}_{text}$ is the text embedding, $t$ is timestep, $\epsilon$ is the noise of the diffusion model, and $\lambda$ is the weight of FAIR.
$\mathbf{M}_r(p)$ is the random face mask which is equal to $\mathbf{M}$ with a $p$ probability and is equal to $1$ in other cases.

\noindent\textbf{Face Condition Drop and Shuffle.}
To increase the variability and the controllability of the portrait, we propose a curriculum learning with face condition drop and shuffle.
Inspired by the Classifier Free Guidance~(CFG)~\cite{cfg}, we train image generation under the guidance of "paired faces", "unpaired identities", and "paired identities + unpaired faces".
Specifically, we conduct random face condition drop and shuffling from the same identity, corresponding to the "unpaired identities" and "paired identities + unpaired faces" respectively, before feeding to the diffusion model, as shown in Fig.~\ref{fig:overview}.
During the training, we adopt curriculum learning by gradually adding the shuffling probability for better convergence.
As such, FACT can maintain the robust and flexible generation ability from the original Stable Diffusion and achieve facial construction through face adapter.
During the inference, the denoising process is formulated as:
\begin{equation}
\begin{aligned}
    \label{eq:cfg}
    \mathbf{z}_{t-1} &= (1 + \lambda_{CFG}) \cdot \mathcal{H}(\mathbf{z}_t, t, \mathbf{e}_{text}, \mathbf{e}_{id})\\
    &- \lambda_{CFG}(\mathbf{z}_t, t, \mathbf{e}_{text-neg}, \mathbf{e}_{id-drop}),
\end{aligned}
\end{equation}
where $\mathcal{H}$ is the forward pass for denoising U-Net, and $\lambda_{CFG}$ is the CFG scale.
We use the same CFG scale and the condition drop probability for both the identity and text conditions for simplification.

%%%%%%%%%%%%%%%%%%%%%%%%%%%%%%%% Methods %%%%%%%%%%%%%%%%%%%%%%%%%%%%%%
\section{Experiment}
\label{sec:experiment}
\subsection{Experimental Setup}

\noindent\textbf{Implementation Details.}
We employ Stable Diffusion v1.5~\cite{sd} architecture as the base diffusion model, and the resolution of training data is resized to $512 \times 512$ pixels correspondingly.
The data preprocessing pipeline is similar to PhotoMaker~\cite{photomaker} for identity-oriented dataset.
The identity feature projection module contains $4$ Transformer blocks to get $\mathbf{e}_{id}$.
Face adapters are inserted into each Transformer block of the diffusion model, in which the scale $\alpha$ is set to $1.0$ and $0.5$ respectively in training and inference.
The CFG scale and condition drop probability for identity and text condition are set to $7.0$ and $0.1$ respectively.
The probability of face condition shuffling increases from $0.2$ to $0.6$ during training.
For the objective function, the probability of masked diffusion loss is set to $0.5$, and the weight of FAIR $\lambda$ is set to $0.01$.
The overall framework is optimized using Adam~\cite{Adam} with learning rate $1e-4$ on 8 NVIDIA V100 GPUs for a week. 
The training batch size is $32$.

\noindent\textbf{Evaluation Metrics.}
We conduct text-to-image generation as well as inpainting-based portrait generation to evaluate the effectiveness of adapter-based personalization methods.
Our evaluation dataset includes 20 IDs collected by ourselves, which did not appear in the training set.
For text-to-image generation, we use the same 40 prompts as PhotoMaker~\cite{photomaker} covering a variety of expressions, attributes, actions and backgrounds.
We use CLIP-T~\cite{clipt} metric to measure the text-to-image alignment.
Considering the difference of base models among the methods, we also compute the drop of CLIP-T compared to the corresponding base model for each method for better evaluating the influence of the proposed adapters.
Following the existing identity-preserved personalization approaches~\cite{photomaker, instantid, caphuman, face0}, we evaluate the CLIP-I~\cite{clipi} score as well as the face similarity using embeddings extracted by FaceNet~\cite{facenet} between the generated image and the reference image to measure the identity preservation.
We employ the FID metric~\cite{fid} to evaluate the quality of the generation.
Besides, we compute the CLIP-I score between the images generated by models with and without adapter for measuring the style consistency when adding the adapter to the diffusion model.
For inpainting-based portrait generation, we prepare 100 portrait images as templates.
Due to the constraint of the facial region in the inpainting process, there is a trade-off between overall harmony and facial shape matching.
In order to measure them, we leverage the mean squared error (MSE) of the DECA (Detailed Expression Capture and Animation)~\cite{DECA} codes, which is commonly used in 3D face modeling.
Specifically, we evaluate the MSE of expression (E-Exp.), lighting (E-Light), and pose (E-Pose) for the generated image and the template image for overall harmony, as well as the MSE of shape (E-Shape) for the generated image and the reference image for face shape matching.
In addition, the FID score is evaluated between the generated results and the templates in this task for better measuring the inpainting quality.

\subsection{Personalization Results}

\begin{figure*}[t]
  \centering
  \includegraphics[width=1.0\linewidth]{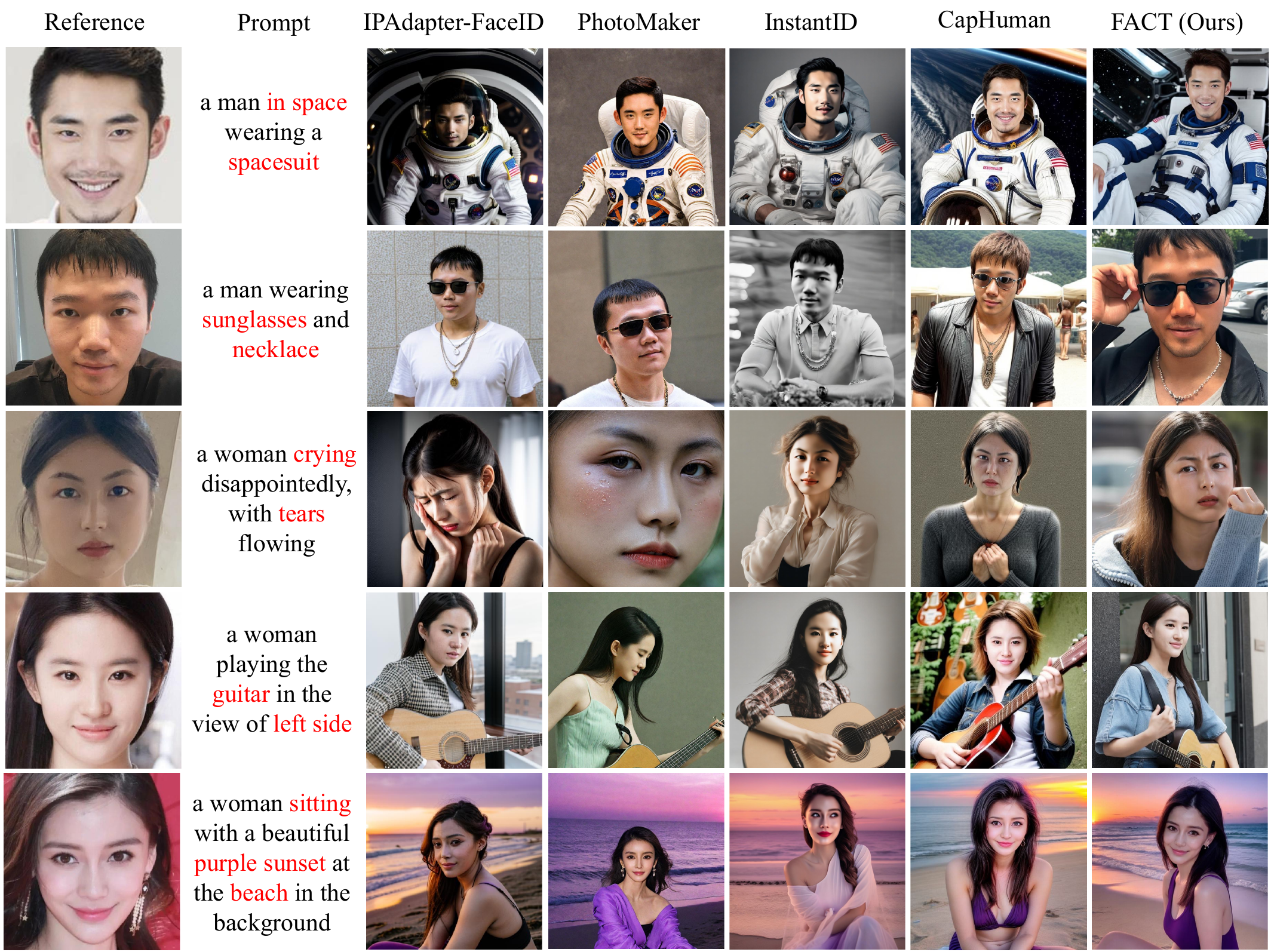}
  \caption{\textbf{Qualitative comparison on text-to-image generation for different adapter-based methods.} Our method generally achieves high identity fidelity, face editability, text following ability, and generation quality.}
  \label{fig:rst_t2i}
\end{figure*}

\noindent\textbf{Text-to-Image Generation.}
We compare our proposed FACT with the newest adapter-based identity-preserved personalization methods including IPAdapter-FaceID~\cite{ipadapter} (simplified as IPAdapter), PhotoMaker~\cite{photomaker}, InstantID~\cite{instantid}, and CapHuman~\cite{caphuman}.
Table~\ref{tab:comp-t2i} shows the quantitative results.
In terms of text-to-image alignment, FACT's CLIP-T score is second only to PhotoMaker.
This is because they used SDXL~\cite{sdxl} as the base model, which has larger parameters than SD v1.5.
When comparing with the corresponding base model, FACT has the lowest drop of CLIP-T, indicating that the text-to-image ability is maintained to the greatest extent by decoupling portrait from overall generation.
In terms of identity preservation, FACT has the highest CLIP-I and the second highest face similarity.
Despite leading the face similarity, InstantID has the largest CLIP-T drop for all comparing methods, and the facial texture is also unsatisfactory, making its CLIP-I not that high, indicating that it may overfit to information beyond identity from the reference face.
In terms of style consistency and image quality, FACT leads in the CLIP Style and FID scores, validating the effectiveness of the proposed decoupled training in maintaining the overall text-to-image ability of the base model.

\begin{table}[t]
  \caption{\textbf{Quantitative comparison on text-to-image generation for different adapter-based methods.} The best result is shown in \textbf{bold}, and the second best is \underline{underlined}.}
  \label{tab:comp-t2i}
  \centering
  \resizebox{1.0\linewidth}{!}{
  \begin{tabular}{lllllll}
    \toprule
    Method  & CLIP-T$\uparrow$ & $\Delta$CLIP-T$\downarrow$ & CLIP-I$\uparrow$ & Face Sim.$\uparrow$ & CLIP Style$\uparrow$ & FID$\downarrow$ \\
    \midrule
    IPAdapter~\cite{ipadapter} & 28.9\% & 1.6\% & 68.3\% & 69.0\% & \underline{73.2}\% & \underline{176.7}\\
    PhotoMaker~\cite{photomaker} & \textbf{30.1}\% & \underline{0.8}\% & \underline{71.7}\% & 70.5\% & 67.3\% & 180.2\\
    InstantID~\cite{instantid} & 27.8\% & 2.1\% & 70.7\% & \textbf{84.2}\% & 69.0\% & 181.0\\
    CapHuman~\cite{caphuman} & 28.5\% & 2.0\% & 70.5\% & 77.2\% & 70.4\% & 186.3\\
    FACT (Ours) & \underline{29.5}\% & \textbf{0.5}\% & \textbf{71.8}\% & \underline{80.6}\% & \textbf{75.2}\% & \textbf{176.5}\\
    \bottomrule
  \end{tabular}
  }
\end{table}

We also show the visual generation results for different reference images and text prompts in Figure~\ref{fig:rst_t2i}.
The image quality and text following ability of IPAdapter are satisfactory, but the face similarity of the generated images is not sufficient, which is also evident in the Face Sim. column of Table~\ref{tab:comp-t2i}.
PhotoMaker has similar issues with IPAdapter, and it also faces other problems, such as the centered position of the head (1st, 2nd, 3rd, and 5th row) and incorrect background (1st row) due to overfitting to the training dataset.
The reason for such unsatisfying facial similarity is that IPAdapter and PhotoMaker only use a global face embedding extracted by a general encoder, resulting in the loss of detailed information about the face.
In contrast, we leverage the fine-grained identity features extracted by a face expert encoder to better preserve the detailed information of the face, thereby enhancing the visual similarity of the generated face.
By introducing extra spatial control, InstantID and CapHuman have more stable face poses and higher face similarity.
However, the parallel network architecture of those methods is harmful to the spatial consistency of the face and body, such as the texture for InstantID and the face color for CapHuman (5th row).
Moreover, the face editability (2nd row) and pose diversity (4th row) are also sacrificed for InstantID and CapHuman.
Different from those methods, FACT effectively decouples the identity feature from the reference image, and achieves high identity fidelity as well as high text following ability, including different clothes, backgrounds, attributes and expressions, as is shown in the right column in Figure~\ref{fig:rst_t2i}.

\begin{figure*}[t]
  \centering
  \includegraphics[width=1.0\linewidth]{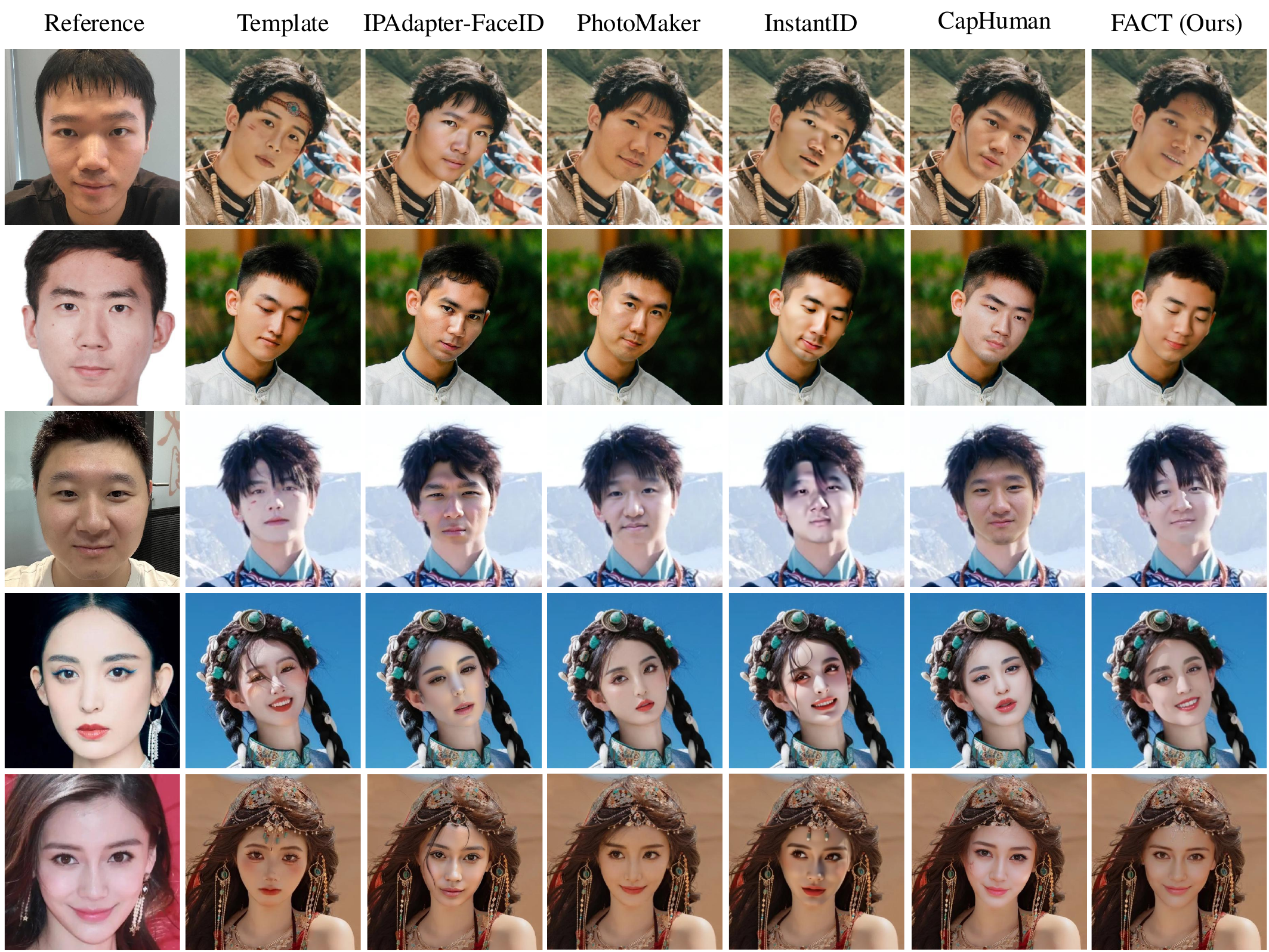}
  \caption{\textbf{Qualitative comparison on inpainting-based portrait generation for different adapter-based methods.} Our method generally achieves high identity fidelity, face shape fitting, and inpainting harmony in expression, light and pose.}
  \label{fig:rst_inpaint}
\end{figure*}

\begin{table}[t]
  \caption{\textbf{Quantitative comparison on inpainting-based portrait generation for different adapter-based methods.} The best result is shown in \textbf{bold}, and the second best is \underline{underlined}.}
  \label{tab:comp-inpaint}
  \centering
  \resizebox{1.0\linewidth}{!}{
  \begin{tabular}{llllllll}
    \toprule
    Method  & E-Exp.$\downarrow$ & E-Light$\downarrow$ & E-Pose$\downarrow$ & E-Shape$\downarrow$ & Face Sim.$\uparrow$ & CLIP-I$\uparrow$ & FID$\downarrow$ \\
    \midrule
    IPAdapter~\cite{ipadapter} & 5.31 & 0.580 & 0.0259 & 6.97 & 70.4\% & 69.4\% & 40.7 \\
    PhotoMaker~\cite{photomaker} & 4.63 & \underline{0.286} & 0.0110 & 7.40 & 69.9\% & \underline{72.8}\% & \underline{30.6} \\
    InstantID~\cite{instantid} & 3.47 & 0.298 & \textbf{0.0043} & 6.87 & \textbf{80.8}\% & 70.4\% & 32.0 \\
    CapHuman~\cite{caphuman} & \underline{3.08} & 0.569 & 0.0103 & \textbf{6.15} & 76.6\% & 72.3\% & 37.9 \\
    FACT (Ours) & \textbf{3.06} & \textbf{0.282} & \underline{0.0057} & \underline{6.26} & \underline{78.2}\% & \textbf{73.3}\% & \textbf{30.4} \\
    \bottomrule
  \end{tabular}
  }
\end{table}

\noindent\textbf{Inpainting-based Portrait Generation.}
We compare FACT with IPAdapter~\cite{ipadapter}, PhotoMaker~\cite{photomaker}, InstantID~\cite{instantid} and CapHuman~\cite{caphuman} and evaluate the identity fidelity, inpainting harmony and face shape matching ability in Table~\ref{tab:comp-inpaint}.
For identity preservation, FACT has the highest CLIP-I and the second highest Face Sim. only lower than InstandID.
However, compared to the sense of replication and disharmonious facial texture in InstantID, our results show better vividness and diversity.
For the trade-off between inpainting harmony and face shape matching, FACT produces lighting and expressions that best align with the template, while offering competitive pose and shape matching capabilities.
It is noteworthy that the leaders for E-Pose and E-Shape are InstantID and CapHuman, which use additional facial landmarks from the template image and the DECA shape code from the reference image respectively.
That implies that their generated results tend to over-align with the pose from the template image and face shape from the reference image, potentially introducing unwanted side effects on the overall inpainting effectiveness.
In other words, FACT exhibits the best matching ability in terms of expression, lighting, pose and shape among all compared methods, unless there is direct additional control for a specific metric.
Moreover, FACT has the best FID score.

We also show the visual generation results in Figure~\ref{fig:rst_inpaint}.
It can be observed that similar problems persist in the comparing methods, including the poor identity similarity exhibited by IPAdapter and PhotoMaker, the texture inconsistency between the face and body in InstantID, and the discrepancy in skin color between the face and the body in CapHuman.
In addition, the trade-off between inpainting harmony and face shape matching can also seen in Figure~\ref{fig:rst_inpaint}.
For example, the pose matching the template image and the face shape matching the reference image may conflict with each other, and fitting one of them like InstantID or CapHuman could compromise the other.
As is shown in the 1st and 5th rows in InstantID's results, the pose adopted from the template image results in a decrease in visual similarity to the reference image in terms of face shape.
The 1st and 3rd rows in CapHuman's results show another case where over-alignment to the face shape of the reference image leads to a disharmonious pose.
Therefore, an ideal result for inpainting-based portrait generation should consider the overall harmony including expression, lighting, pose, face shape and identity similarity.
From this perspective, FACT significantly outperforms the comparing methods with the help of decoupled training.
Different from other methods which aim to denoise towards the reference image in both training and inference for face adapter, FACT modifies the original denoising process of the template image by only adjusting the identity to that in the reference image, thus achieving the most harmonious inpainting results while maintaining high identity fidelity.

\begin{figure*}[t]
  \centering
  \includegraphics[width=1.0\linewidth]{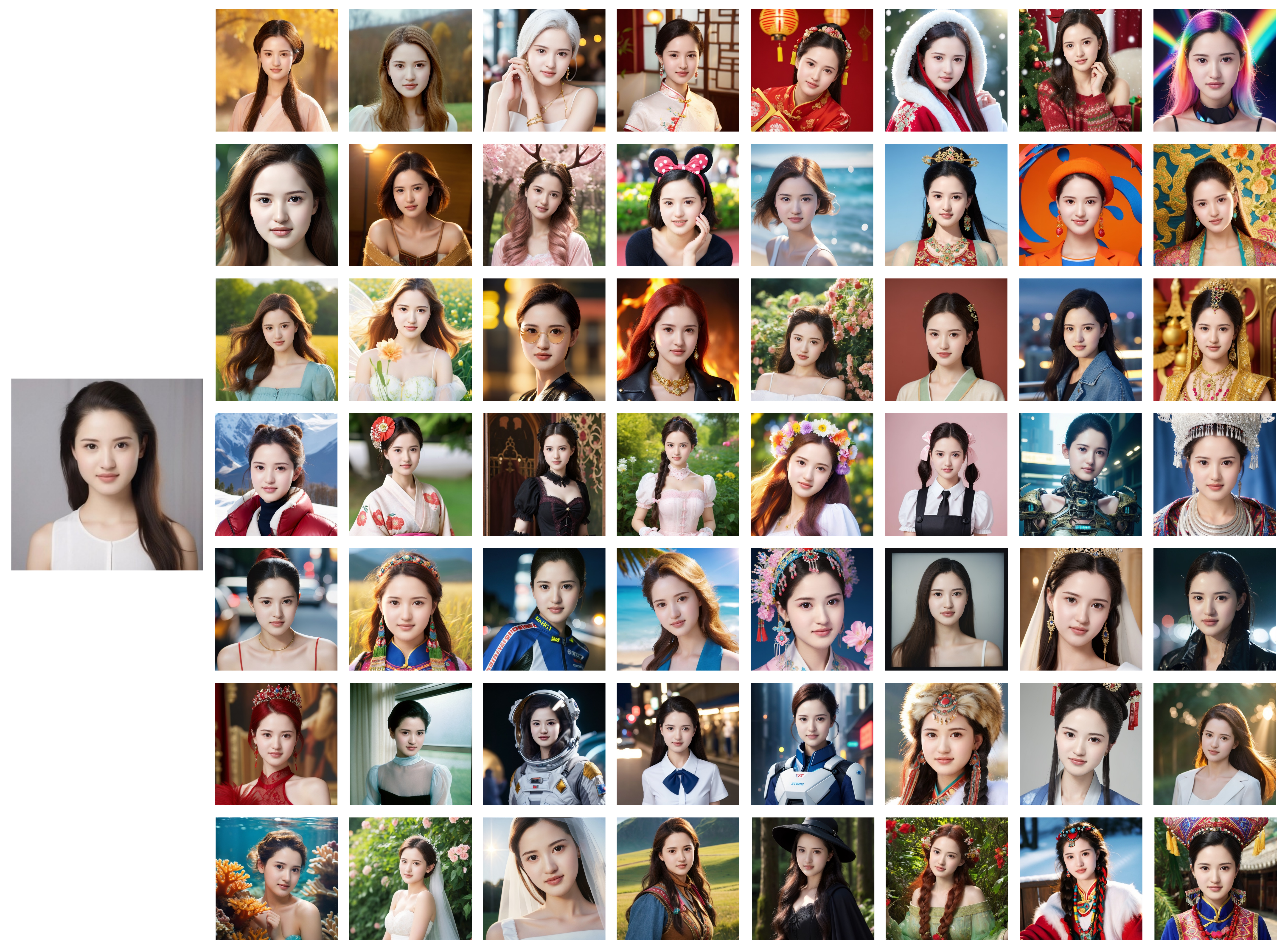}
  \caption{\textbf{Generated results with various style-LoRA models by FACT.}}
  \label{fig:rst_lora}
\end{figure*}

\begin{figure*}[t]
  \centering
  \includegraphics[width=1.0\linewidth]{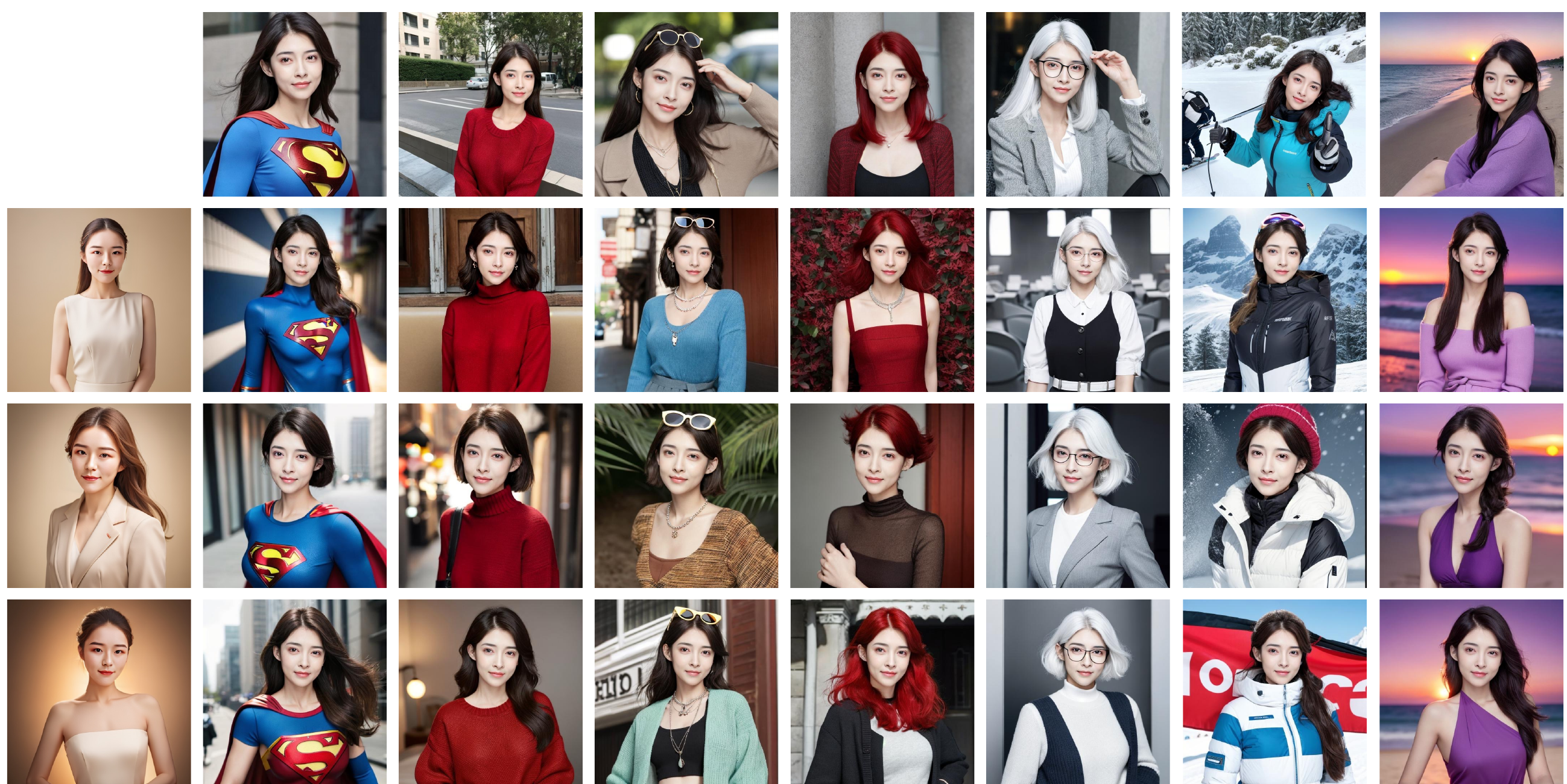}
  \caption{\textbf{Generated results using OpenPose~\cite{openpose} ControlNet~\cite{controlnet} with different text prompts.}}
  \label{fig:rst_controlnet}
\end{figure*}

\noindent\textbf{Combining with LoRA and ControlNet.}
In real-world human portrait generation, there is a very important requirement for accurately controlling contents like clothing, background, and pose information to simulate photography operations.
This requires the model to have higher controllability and style consistency than the base Stable Diffusion.
To achieve this, common fine-tuning approaches like LoRA~\cite{lora} and ControlNet~\cite{controlnet} are proposed to control the style and pose of the portrait.
In this section, we demonstrate the compatibility of our proposed FACT with LoRA and ControlNet to verify the generalization ability of FACT.

Figure~\ref{fig:rst_lora} shows the generated results of FACT combining with various style-LoRA models from FaceChain~\cite{facechain}.
Due to the decoupling strategy, FACT focuses on adapting the identity information to the input face, and retains the detailed and high-quality style information of the style LoRAs as much as possible.
As such, the generated images achieve accurate identity preservation as well as rich and detailed portrait styles without mutual interference.
Figure~\ref{fig:rst_controlnet} shows the generated results of FACT combined with OpenPose~\cite{openpose} ControlNet using different human poses and text prompts.
For each row, the pose of the generated portraits keeps same as the reference image.
For each column, portraits with the same text prompt have high style consistency and certain diversity.
Moreover, all generated portraits preserve the high image quality of the base model, validating the high compatibility and generalization ability of FACT.

\subsection{Ablation Study}

\noindent\textbf{Effect of Face Adapting Increment Regularization.}
We evaluate the effect of our proposed FAIR by comparing the text-to-image generation quality for FACT with and w without (w/o) FAIR.
Table~\ref{tab:abl1} shows the comparison of metrics on identity preservation (CLIP-I, Face Sim.) and text-to-image generation ability (CLIP-T, CLIP Style), which is kept in Ablation Study.
It can be observed that our proposed FAIR has a positive effect on all metrics.
FAIR helps the face adapter to focus on the face region and perform identity preservation on it, while maintaining the attention on different areas of the face on different depths of Transformer blocks, which is demonstrated in Figure~\ref{fig:fair}.
As a result, the face adapter can concentrate solely on identity preservation, while the base model can focus entirely on text-to-image generation, without interfering with each other.
This leads to an improvement in the metrics on both roles.

\begin{table}[h]
  \caption{\textbf{Comparison on text-to-image generation for models with and w/o FAIR.} The best result is shown in \textbf{bold}.}
  \label{tab:abl1}
  \centering
  \resizebox{1.0\linewidth}{!}{
  \begin{tabular}{lllll}
    \toprule
    Method  & CLIP-T$\uparrow$ & CLIP-I$\uparrow$ & Face Sim.$\uparrow$ & CLIP Style$\uparrow$\\
    \midrule
    w/o FAIR & 29.3\% & 71.6\% & 80.3\% & 74.2\% \\
    with FAIR & \textbf{29.5}\% & \textbf{71.8}\% & \textbf{80.6}\% & \textbf{75.2}\%\\
    \bottomrule
  \end{tabular}
  }
\end{table}

\noindent\textbf{Effect of Face Condition Drop and Shuffle.}
We perform text-to-image generation for models with different face condition drop and shuffle strategies, including model without face condition drop and shuffle (w/o DS), model with fixed drop and shuffle probability (w/o CL), and model with face condition drop and shuffle by curriculum learning (DS + CL).
Table~\ref{tab:abl2} shows the comparison result.
Comparing the first two lines, we find that both CLIP-T and face similarity metrics benefit from face condition drop and shuffle despite a slight drop in CLIP Style.
The increase of CLIP-T comes from the improvement in editability of generated faces by face condition shuffle, and the increase of face similarity metrics is due to the classifier-free guidance of face condition with the help of face condition drop.
The comparison between the last two rows of data reveals that the model incorporating curriculum learning demonstrates superior identity preservation capabilities while maintaining the overall text-to-image generation quality, thereby validating the effectiveness of the proposed curriculum learning strategy for face condition drop and shuffle.

\begin{table}[h]
  \caption{\textbf{Comparison on text-to-image generation for models with different face condition drop and shuffle strategies.} The best result is shown in \textbf{bold}.}
  \label{tab:abl2}
  \centering
  \resizebox{1.0\linewidth}{!}{
  \begin{tabular}{lllll}
    \toprule
    Method  & CLIP-T$\uparrow$ & CLIP-I$\uparrow$ & Face Sim.$\uparrow$ & CLIP Style$\uparrow$\\
    \midrule
    w/o DS & 29.1\% & 70.0\% & 76.9\% & \textbf{75.8}\% \\
    w/o CL & \textbf{29.5}\% & 71.3\% & 79.5\% & 75.4\%\\
    DS + CL & \textbf{29.5}\% & \textbf{71.8}\% & \textbf{80.6}\% & 75.2\%\\
    \bottomrule
  \end{tabular}
  }
\end{table}

\begin{figure}[t]
  \centering
  \includegraphics[width=1.0\linewidth]{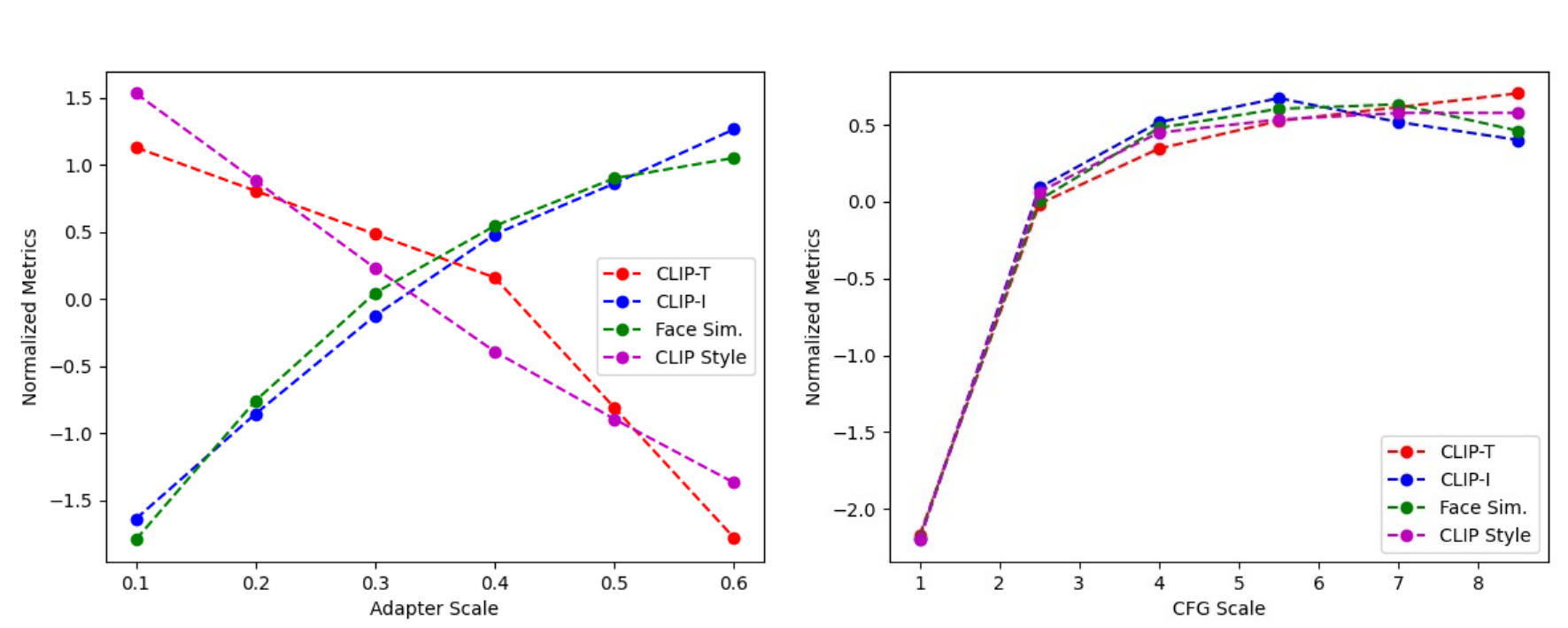}
  \caption{\textbf{Comparison on text-to-image generation for different adapter scales and CFG scales.} All metrics are normalized by their mean values and standard variations.}
  \label{fig:abl3}
\end{figure}

\noindent\textbf{Influence of Inference Strategies.}
We illustrate the curve of metrics on text-to-image generation by changing the adapter scale $\alpha$ and CFG scale in Figure~\ref{fig:abl3}.
It can be observed that there is a trade-off between the identity preservation and the text following ability when changing $\alpha$.
In our experiments, we choose $\alpha=0.5$ to balance the metrics.
When the CFG scale is relatively small (less than $4.0$), higher CFG scale leads to higher overall performance.
Then, the overall performance is stable when the CFG scale is between $4.0$ and $7.0$.
When the CFG scale is too high, there may exist overflow in the generated faces, leading to the performance drop of the identity preservation.

%%%%%%%%%%%%%%%%%%%%%%%%%%%%%%%% Methods %%%%%%%%%%%%%%%%%%%%%%%%%%%%%%
\section{Conclusion}
\label{sec:conclusion}
We propose the Face Adapter with deCoupled Training (FACT), an identity-preserved personalization text-to-image generation method.
We leverage a transformer-based face-export encoder and harness fine-grained identity features to decouple identity features from others.
Furthermore, we insert an adapter with a sequential Gated Self-Attention (GSA) to adapt independently to facial features, minimizing the interference from textual embeddings.
To decouple the portrait generation training, we propose Face Adapting Increment Regularization~(FAIR), which effectively constrains the effect of face adapters on the facial region, preserving the ability of the original model.
With the help of the above decoupling strategy, FACT performs personalization solely by learning identity preservation from training data, thereby minimizing the performance impact on the original text-to-image capabilities of the base model.
Extensive Experiments demonstrate that FACT exhibits both controllability and fidelity in both text-to-image generation and inpainting solutions for portrait generation.

%%%%%%%%% REFERENCES
{\small
\bibliographystyle{ieee_fullname}
\bibliography{references}
}

\end{document}